\documentclass[runningheads]{llncs}
\usepackage[T1]{fontenc}
\usepackage{subcaption}
\usepackage{graphicx}
\usepackage{booktabs}
\usepackage{multirow}
\usepackage{tabu}
\usepackage{longtable}
\usepackage{rotating} 
\usepackage{diagbox}
\usepackage{xcolor}
\usepackage{colortbl}
\definecolor{lightgray}{gray}{0.9}
\usepackage{xspace}
\usepackage{wrapfig}
\usepackage{cite}
\usepackage[accsupp]{axessibility}
\usepackage[pagebackref,breaklinks,colorlinks,citecolor=blue]{hyperref}
\usepackage{orcidlink}

\newcolumntype{L}[1]{>{\raggedright\arraybackslash}p{#1}}
\newcolumntype{R}[1]{>{\raggedright\arraybackslash}p{#1}}

\newcommand{\repthanks}[1]{\textsuperscript{\ref{#1}}}
\makeatletter
\patchcmd{\maketitle}
{\def\thanks}
{\let\repthanks\repthanksunskip\def\thanks}
{}{}
\patchcmd{\@maketitle}
{\def\thanks}
{\let\repthanks\@gobble\def\thanks}
{}{}
\newcommand\repthanksunskip[1]{\unskip{}}
\makeatother

\makeatletter
\DeclareRobustCommand\onedot{\futurelet\@let@token\@onedot}
\def\@onedot{\ifx\@let@token.\else.\null\fi\xspace}

\def\eg{\emph{e.g}\onedot} 
\def\ie{\emph{i.e}\onedot}

\begin{document}

\title{Automated Image Recognition Framework} 

\titlerunning{Automated Image Recognition Framework}

\author{
Quang-Binh Nguyen\thanks{These authors contributed equally to this work.\protect\label{X}}\inst{1,2}\orcidlink{0000-0003-1199-3661} \and
Trong-Vu Hoang\repthanks{X}\inst{1,2}\orcidlink{0000-0001-7367-1401} \and
Ngoc-Do Tran\inst{1,2}\orcidlink{0009-0002-2375-342X} \and \\
Tam V. Nguyen\inst{3}\orcidlink{0000-0003-0236-7992} \and
Minh-Triet Tran\inst{1,2}\orcidlink{0000-0003-3046-3041} \and
Trung-Nghia Le\thanks{Corresponding author. Email: ltnghia@fit.hcmus.edu.vn}\inst{1,2}\orcidlink{0000-0002-7363-2610} 
}

\authorrunning{Q.-B. Nguyen et al.}

\institute{University of Science, VNU-HCM, Ho Chi Minh City, Vietnam \and
Vietnam National University, Ho Chi Minh City, Vietnam \and
University of Dayton, Ohio, United States 
}

\maketitle

\begin{abstract}
  While the efficacy of deep learning models heavily relies on data, gathering and annotating data for specific tasks, particularly when addressing novel or sensitive subjects lacking relevant datasets, poses significant time and resource challenges. In response to this, we propose a novel Automated Image Recognition (AIR) framework that harnesses the power of generative AI. AIR empowers end-users to synthesize high-quality, pre-annotated datasets, eliminating the necessity for manual labeling. It also automatically trains deep learning models on the generated datasets with robust image recognition performance. Our framework includes two main data synthesis processes, AIR-Gen and AIR-Aug. The AIR-Gen enables end-users to seamlessly generate datasets tailored to their specifications. To improve image quality, we introduce a novel automated prompt engineering module that leverages the capabilities of large language models. We also introduce a distribution adjustment algorithm to eliminate duplicates and outliers, enhancing the robustness and reliability of generated datasets. On the other hand, the AIR-Aug enhances a given dataset, thereby improving the performance of deep classifier models. AIR-Aug is particularly beneficial when users have limited data for specific tasks. Through comprehensive experiments, we demonstrated the efficacy of our generated data in training deep learning models and showcased the system's potential to provide image recognition models for a wide range of objects. We also conducted a user study that achieved an impressive score of 4.4 out of 5.0, underscoring the AI community's positive perception of AIR.
  \keywords{Generative AI \and Image Recognition \and Dataset Generation \and Data Augmentation}
\end{abstract}

\section{Introduction}


In the era of AI, particularly within deep learning research, data plays a crucial role, serving as a pivotal factor in both the academic and industrial arenas. Nevertheless, collecting and annotating data for specific tasks frequently demands extensive time and effort~\cite{ltnghia-WACV2020}, especially when dealing with new concepts or sensitive subjects lacking relevant existing datasets. For example, we consider a task called \textit{early forest fire recognition}. Collecting a sufficient quantity of high-quality images of early forest fires for training deep learning models presents a significant challenge. Most images available on the Internet depict fires that have already spread widely. Additionally, creating our own dataset by starting small fires in forest and capturing images carries potential risks, time-consuming and often does not accurately represent real world situations. Therefore, it is essential to develop a system capable of generating pre-annotated datasets tailored to user specifications, producing images of a sufficiently high level of realism to be applicable in real-world scenarios without manual human labeling costs.

As Generative AI continues to advance significantly in both computer vision and natural language processing fields~~\cite{Goodfellow-GAN2020, Rombach-CVPR2022, Radford-GPT2018}, we acknowledge its potential in building solutions for generating datasets customized to the user's specific demands. Regarding image generation, numerous robust models have emerged, capable of generating images based on textual descriptions~\cite{
kang2023scaling
}. Notably, diffusion models~\cite{Rombach-CVPR2022 
} have garnered attention for their ability to produce images of remarkable quality and realism. 

\begin{figure}[t!]
	\centering
 \includegraphics[height=6.5cm]{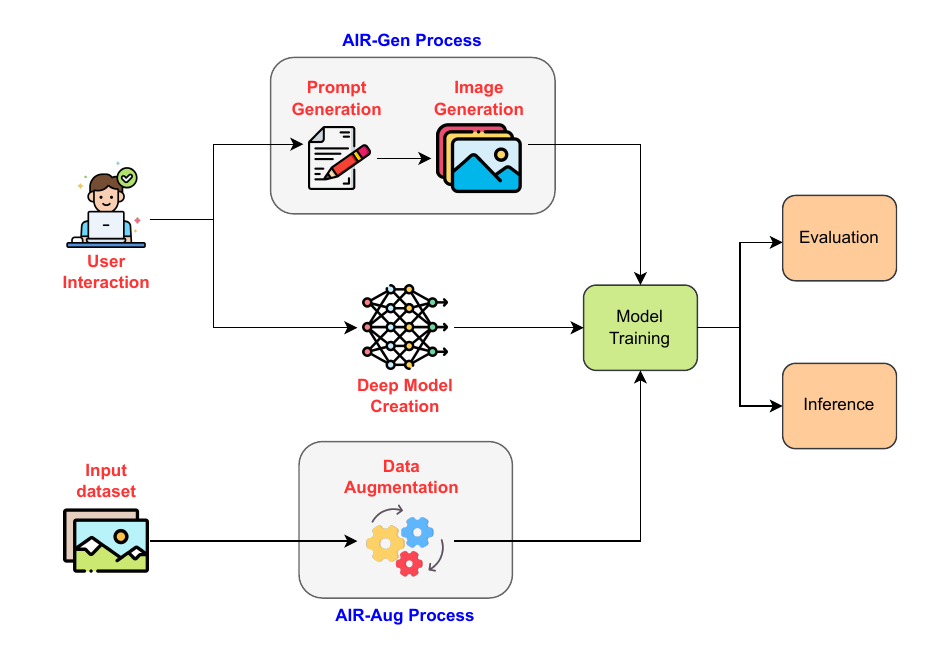}
        \vspace{-5mm}
	\caption{Overall pipeline of our proposed AIR framework with two primary processes for data synthesis: AIR-Gen creating new datasets and AIR-Aug generating augmented data for a given dataset.}
	\label{fig:overall_system}
 \vspace{-5mm}
\end{figure}

With these advancements in mind, we harness cutting-edge generative AI technologies to introduce a novel system called \textbf{A}utomated \textbf{I}mage \textbf{R}ecognition framework (AIR), comprising two primary data synthesis processes: AIR Generation (AIR-Gen) and AIR Augmentation (AIR-Aug) (see Fig.~\ref{fig:overall_system}). 
AIR is specifically designed to cater to a wide spectrum of end-users to train deep models for image classification.
Even without extensive AI expertise as well as available appropriate datasets, users can simply describe what they seek to recognize. The AIR-Gen module subsequently generates realistic image datasets and then autonomously trains deep classifiers, empowering users to leverage trained models for predictions on their own images. Meanwhile, AIR-Aug augments a given dataset, thereby boosting the performance of deep classifier models, which is especially advantageous when users have limited data for specific tasks.

While text-to-image diffusion models~\cite{
saharia2022photorealistic, balaji2022ediffi, nichol2021glide} can create impressive images from textual descriptions, the formulation of descriptive sentences with specific structures can significantly affect to the quality of the generated images. To address this issue, we employ in-context learning techniques in conjunction with powerful large language models, such as GPT-3.5-Turbo
, to develop an automated prompt engineering module. This novel module streamlines the fine-tuning of input prompts for text-to-image models, resulting in the production of higher-quality, realistic images in the module AIR-Gen. 

In the data augmentation process of AIR-Aug, we focus on creating a simulated dataset that captures the essential characteristics of the original dataset, thereby maintaining the dataset's diversity and providing a contextually accurate environment for model training. Particularly, we directly extract textual prompts from the images in the given dataset using the CLIP~\cite{clip} text decoder and use these prompts for the image generation process. Furthermore, we incorporate a style transfer step optionally by training CycleGAN~\cite{cyclegan} model to help lessen the difference in image quality and visual attributes (\eg, noise, blurriness, brightness, and color temperature) between the provided dataset and the images created by the text-to-image model.

To demonstrate the real-world utility of our system, extensive experiments are conducted to evaluate the usability of generated datasets. The efficacy of AIR-Gen in the context of early forest fire recognition is evaluated, resulting in a dataset exhibiting an appropriate distribution compared to real images on the Internet, leading to an impressive accuracy of 95.48\% in training image classification models. 
In addition, to assess the capacity of AIR-Aug in providing augmented images to improve the performance of image recognition models, we augment the MIT Indoor Scene dataset~\cite{quattoni2009recognizing} and integrate it with the original dataset for training a deep classifier model. The experimental findings reveal the potential of AIR-Aug in enhancing the training of deep models when authentic data is limited, resulting in a boost in recognition accuracy of approximately 4\%. Furthermore, to gauge the effectiveness of our method, we conduct a user study by letting participants experience AIR, especially AIR-Gen. The outcome of our survey demonstrates a high overall user satisfaction, resulting in an impressive overall score of 4.4 out of 5.0. This reflects exceedingly positive user experiences with our AIR. In summary, our main contributions are as follows:

\begin{itemize}
    \item We introduce the novel AIR framework, including AIR-Gen, an approach that empowers users to effortlessly create custom datasets tailored to their specific requirements, and AIR-Aug, a data augmentation method designed to further enhance limited datasets in real-world scenarios. This obviates the extensive human labeling process and facilitates the automatic training of deep learning models.

    \item We propose a novel automated prompt engineering module, simplifying the process of generating well-designed prompts for text-to-image tasks.

    \item We showcase the real-world applicability of our generated data in deep learning model training, highlighting the effectiveness of our method in providing deep models for a wide range of image recognition tasks.
    
\end{itemize}

\section{Related Work}

Generative model-based text-to-image, driven by large-scale datasets 
, is rapidly evolving, drawing widespread attention from scientists and the public alike. Notably, these advancements center around two dominant approaches: GANs~\cite{Goodfellow-GAN2020, cyclegan
} known for high-fidelity image synthesis and fast inference but contend with training instability and diversity issues, diffusion models~\cite{
saharia2022photorealistic, balaji2022ediffi, nichol2021glide} excelling in producing high-quality, realistic images. Pioneering works~\cite{nichol2021glide
} integrated diffusion model with classifier guidance to improve image quality. After that, to balance efficiency and fidelity, novel techniques have emerged, such as the coarse-to-fine generation~\cite{saharia2022photorealistic, balaji2022ediffi} and latent space exploration~\cite{Rombach-CVPR2022}. Built on the latent diffusion concept, Stable Diffusion~\cite{Rombach-CVPR2022} 
became a pivotal open-source model, expanding text-to-image synthesis capabilities. Our system adopts Stable Diffusion for the text-to-image generation module, drawn to its remarkable image quality and the availability of robust pre-trained weights. 

Image recognition, a fundamental task in computer vision, involves assigning input images to predefined classes, enabling machines to interpret visuals and make informed decisions, serving as a linchpin for AI applications. In recent years, the field of image recognition has witnessed a transformative revolution driven by the emergence of deep learning models~\cite{
he2016deep, tan2021efficientnetv2, visiontransformer, liu2021swin
}
. These models have demonstrated remarkable capabilities in learning intricate features and patterns from large datasets, substantially improving our capacity to recognize and categorize objects, scenes, and patterns within images. Our system introduces a novel perspective by combining text-to-image generation with image recognition model training. This integration enables users to create customized datasets for training image classifiers, meeting the demand for high-quality, domain-specific data crucial for real-world AI performance. Our system also supports various standout backbone architectures~\cite{he2016deep, tan2021efficientnetv2, visiontransformer, liu2021swin}, enhancing the flexibility and adaptability of the image recognition pipeline, enabling users to harness the potential of their data for precise and context-aware image recognition.

\section{Proposed Method}

\begin{figure*}[t!]
	\centering
 \includegraphics[width=\textwidth]{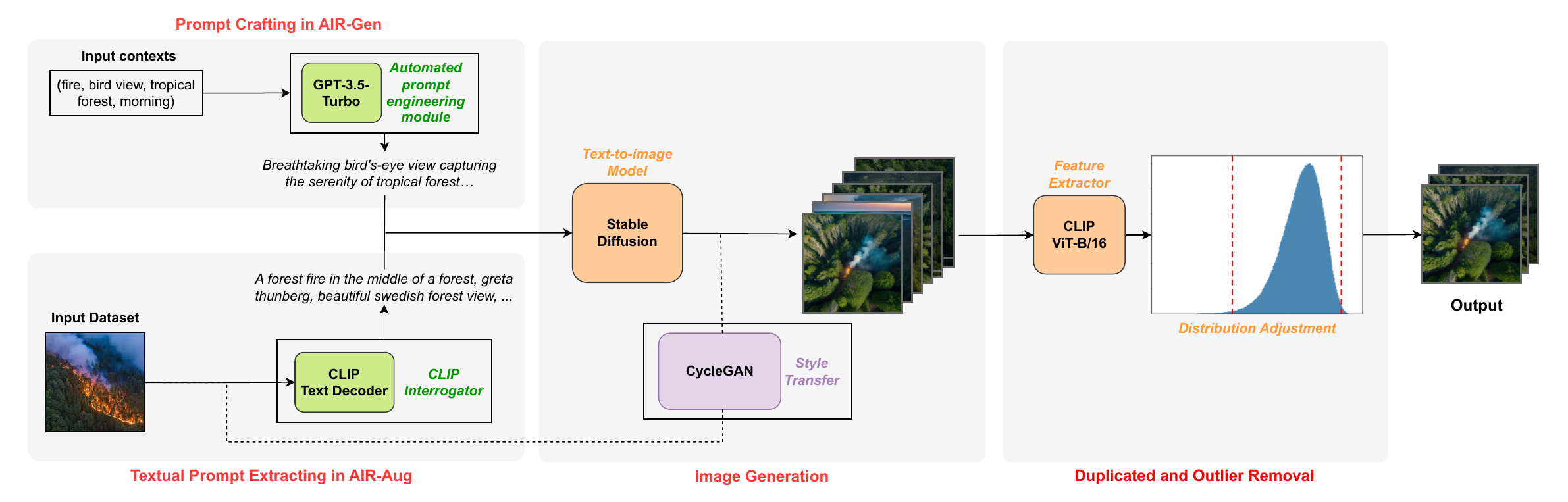}
    \vspace{-5mm}
	\caption{The dataset synthesis process consists of three key phases: prompt generation, image generation, and duplicated and outlier removal.
 The two processes AIR-Gen and AIR-Aug share the two latter phases while differing in the prompt generation phase.
 Also, in AIR-Aug, we include an optional style transfer step to minimize disparities in image quality and visual attributes between the provided dataset and images generated by the text-to-image model.
 }
	\label{fig:overall_data_gen}
 \vspace{-3mm}
\end{figure*}

\subsection{Dataset Synthesis Pipeline}

The dataset synthesis processes of AIR-Gen and AIR-Aug are illustrated in Fig.~\ref{fig:overall_data_gen} with three main phases: prompt generation, image generation, and duplicated and outlier removal.
For AIR-Gen, first of all, the prompt generation phase processes the user-provided description and transforms it to relevant prompts which helps a text-to-image model to generate corresponding images in the image generation phase. On the other hand, the prompt generation phase for AIR-Aug involves extracting textual prompts directly from the original given dataset. 
These prompts are then passed to the image generation phase.
We also incorporate a style transfer phase in AIR-Aug as an optional choice to minimize the disparity in image quality (\eg, noise, blurriness, brightness, and color temperature) between the images in the provided dataset and those generated by the text-to-image model.
Finally, to ensure the quality of the generated dataset in both modules, the duplicated and outlier removal phase filters out all the unqualified images. The user interface for this pipeline is presented in the supplementary.



\subsubsection{Prompt Crafting in AIR-Gen}

The quality and realisticness of images generated by the text-to-image model, particularly Stable Diffusion, relies heavily on the input prompts. Recently, the Civitai community 
~\footnote{\url{https://civitai.com/}} 
demonstrated that highly detailed prompts, including terms like ``4K" and ``high resolution", significantly enhance image quality. They also suggested using terms within parentheses to guide the model's attention (\eg (word:1.4) indicating an attention weight of 140\%. However, crafting effective prompts has traditionally been labor-intensive for users).

To tackle this challenge, we introduce an automated prompt engineering module, alleviating the burden of manual prompt crafting. Users now need only provide essential keywords describing the dataset's context and content. Instead of supervised fine-tuning a pre-trained model on engineered prompts, we adopt an in-context learning approach by providing task descriptions and examples to GPT-3.5-Turbo
. Subsequently, the model adeptly transforms context keywords into well-designed prompts, enhancing prompt quality while minimizing user effort. The template for prompt generation is provided in the supplementary.

\begin{table}[t!]
  \centering
  \caption{Example of input context options and completed prompts generated for the task of early forest fire recognition.}
  \resizebox{0.75\linewidth}{!}{
    \begin{tabular}{cp{8.5cm}}
    \toprule
    \textbf{Context} & \textbf{Input options} \\
    \midrule
    Category & small fire and smoke, normal \\
    Location &  tropical forest, boreal forest\\
    View  & drone's view \\
    Time & morning \\
    \midrule
    \textbf{ID} & \textbf{Generated well-designed prompt} \\
    \midrule
    01    & A captivating drone's view of a tropical forest at dawn, with a medium-intensity fire burning amidst the lush greenery, (dramatic:1.4), (serene:1.2), (vivid:1.3), (medium-intensity fire:1.4), 4K UHD image, Photorealistic, Intricate details\\
    \midrule
    02    & A mesmerizing aerial perspective of a boreal forest at sunrise, featuring a moderately intense fire blazing amidst the vibrant foliage, (captivating:1.4), (serense:1.3), 4K UHD image, Cinematic, Realistic\\
    \midrule
    03    & An evocative drone's view of a tropical forest at sunrise, (mysterious:1.3), (captivating:1.4), (intense:1.2), (lush:1.3),  4K UHD image, Realistic, Fine-tuned, Moody\\
    \midrule
    04    & An enchanting top-down perspective of a temperate forest in the morning, (serene:1.3), (captivating:1.4), (intricate:1.3), 8K resolution, Cinematic, Fine-tuned, Intricate details\\
    \bottomrule
    \end{tabular}
    }
  \label{tab:prompt_example}
  \vspace{-3mm}
\end{table}

To illustrate the process, we walk through the steps using the task of early forest fire recognition as an illustrative example in Table~\ref{tab:prompt_example}:

\begin{itemize}
    \item \textit{Context options:} To generate prompts related to desirable datasets, we define a set of contexts and their corresponding options, which we request from users. The ``category" context is mandatory, determining the dataset's class label. Additional contexts, such as ``location'', ``view captured'', and ``times'' are also provided, each with its respective options.


     \item \textit{Context combinations:} From the provided context options, we create context combinations, each comprising one choice from each context. 


\item \textit{Context weights:} The order of contexts in a combination affects prompt quality and image generation. The earlier a context appears, the more it influences the prompt. This order can be considered a weight for the combination.


\end{itemize}

Afterward, each combination is converted into a well-designed prompt. This process repeats for all combinations, resulting in a set of prompts, as shown in Table~\ref{tab:prompt_example}.

\subsubsection{Extracting Textual Prompts in AIR-Aug}

For this module, our approach centers on the creation of a simulated dataset that effectively encapsulates the core characteristics of the given dataset.

In ensuring the contextual fidelity of each image to its class, our methodology involves using the CLIP~\cite{clip} text decoder to extract textual prompts directly from the given dataset, which is based on CLIP Interrogator technique~\footnote{\url{https://github.com/pharmapsychotic/clip-interrogator}}. Furthermore, it is important to highlight that we do not utilize the automated prompt engineering module for the prompts extracted in AIR-Aug.
These steps are deliberately chosen to avoid the unintended generalization of context across all classes that may occur with auto-generated prompts from our framework. 
By using extracted prompts derived from the original dataset, we maintain the dataset's diversity and offer a fair, contextually accurate environment for model training.


To this end, we extract prompts from a given image dataset, ensuring that prompts correspond to the same number of images. These prompts are then utilized as inputs to the Stable Diffusion model. By leveraging these prompts, the proposed method generates replicated images that closely resemble those found within the original dataset.

\subsubsection{Image Generation}

Generated prompts are fed into Stable Diffusion, a cutting-edge text-to-image model, to produce relevant images. The growing interest in generative AI models has led to the emergence of powerful checkpoints for impressive image generation using Stable Diffusion in various domains. Among these, we specifically utilize the Realistic Vision checkpoint~\footnote{\url{https://civitai.com/models/4201/realistic-vision-v51}} due to its ability to produce exceptionally strikingly realistic images. Additionally, we employ techniques like CPU offloading and attention layer acceleration\footnote{\url{https://github.com/facebookresearch/xformers}} 
to optimize memory and time consumption during the image generation process.

\subsubsection{Style Transfer in AIR-Aug}

By employing text prompts extracted from the input dataset, the AIR-Aug module facilitates the generation of images that align with the dataset's content. However, certain aspects of image quality and visual characteristics (\eg, noise, blurriness, brightness, and color temperature), which we refer to as ``style'', may not be captured by textual prompts and are difficult to faithfully recreate with a pre-trained generative model (see the first two rows in Fig.~\ref{fig:simulated-dataset}).
Hence, we integrate a style transfer stage into AIR-Aug to bridge the stylistic disparity between images in the provided dataset and those generated by the Stable Diffusion model.

Specifically, the generated images are merged with the provided image dataset to train a style transfer model. This model is then utilized to transform the generated images into the same stylistic domain as the provided images. 
Furthermore, given the unpaired structure of the data for training, we employ CycleGAN~\cite{cyclegan}, a well-known and strong model within the realm of unpaired image-to-image translation. 
The output of this phase can be seen in the last row in Fig.~\ref{fig:simulated-dataset}.
It is important to highlight that the training process for CycleGAN can be time-consuming. As a result, we make the style transfer phase in the AIR-Aug module an optional choice.




\begin{figure}[t!]
    \centering
     \begin{subfigure}[t]{0.7\linewidth}
         \centering
         \includegraphics[width=\linewidth]{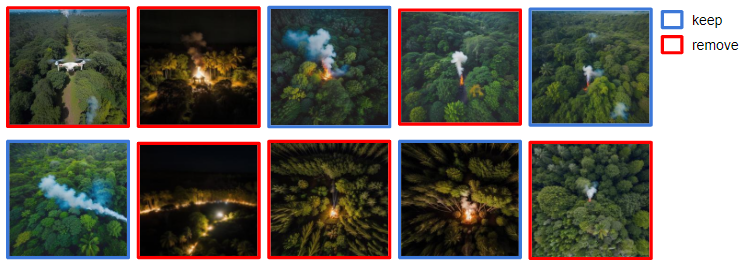}
         \caption{Before}
         \label{fig:before}
     \end{subfigure}
     \hfill
     \begin{subfigure}[t]{0.25\linewidth}
         \centering
         \includegraphics[width=\linewidth]{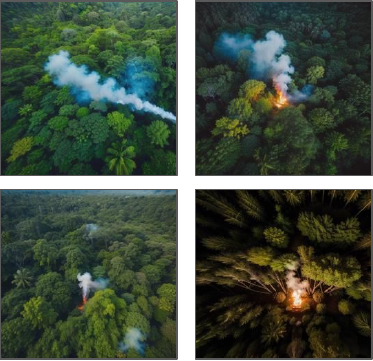}
         
         \caption{After}
         \label{fig:after}
     \end{subfigure}
     \vspace{-3mm}
    \caption{Effectiveness of duplicated and outlier removal.}
    \label{fig:effectiveness_removal}
\vspace{-5mm}
\end{figure}

\subsubsection{Duplicated and Outlier Removal}






Upon examining the generated images stemming from the same prompt, we observe variations in their resemblance to one another. This discovery instigates a post-processing phase (\ie, duplicated and outlier removal) aimed at ensuring the coherence among the images within the dataset. This process is crucial for enhancing the accuracy and reliability of subsequent data-driven models and is achieved through a dataset distribution adjustment operation (see Fig.~\ref{fig:overall_data_gen}).

To facilitate the distribution adjustment process, we compute feature vectors for the generated images using CLIP~\cite{clip}, a robust vision-language model, specifically the B-L/16 version. Each generated image is associated with a corresponding feature vector comprising 512 dimensions. Afterward, in the distribution adjustment process, we assess the image similarity of every image pair using the cosine similarity metric between their extracted features. Let $\mathbf{u}_i$ be the feature vector of image $i$. The similarity score between a pair of images $i$ and $j$ is denoted by $\text{sim}(\mathbf{u}_i, \mathbf{u}_j) = \frac{\mathbf{u}_i \cdot \mathbf{u}_j}{\left \| \mathbf{u}_i \right \| \left \| \mathbf{u}_j \right \|} $. We also define the number of neighbors of an image $i$ as the number of image $j$ ($j \neq i$) by considering their similarity scores in the range $[\alpha, \beta]$.

Our aim is to ensure that the similarity score between any pair of generated images falls within the acceptable range $[\alpha, \beta]$. If a pair fails to meet this condition, the image with fewer neighbors is eliminated. This process effectively eliminates duplicates (whose scores are greater than $\beta$) and irrelevant images (whose scores are less than $\alpha$), illustrated in Fig.~\ref{fig:effectiveness_removal}.
We empirically set $\beta=0.9825$ and finding $\alpha$ such that the retaining ratio is $0.9$.

\subsection{Deep Model Management}

In addition to the dataset generation service, our AIR offers a dedicated platform for deep learning model training, evaluation, and inference. 
This comprehensive service is particularly advantageous for end-users who may lack in-depth expertise in the intricacies of the training process. 
Our platform streamlines and simplifies the creation of custom high-accuracy deep image classification models, ensuring a quick and user-friendly experience tailored for specific purposes.
We offer a range of backbone architectures, spanning from lightweight options to state-of-the-art (SOTA) designs, such as ResNet~\cite{he2016deep}, EfficientNetV2~\cite{tan2021efficientnetv2}, Vision Transformer ~\cite{visiontransformer}, Swin Transformer~\cite{liu2021swin}, and more. 
The backbone architecture chosen by users serves as the feature extraction module for the model. Subsequently, users can personalize their models by adding fully connected layers following the feature extraction module. 



Once the model is created, it becomes ready for training using the generated dataset. 
Users can tailor various training configurations, encompassing data ratio for training and validation, random seed for result reproducibility, optimizer selection, learning rate adjustment, and the number of training epochs. Furthermore, our platform features a visual interface that offers real-time monitoring and analytical capabilities throughout the training phase (see the user interface in the supplementary). This tool empowers users to track model training progress, visualize learning curves, and make informed decisions to optimize training outcomes. 
Upon the completion of the training process, our platform provides valuable insights into model performance, such as training accuracy, validation accuracy, and a confusion matrix on the validation set. 


The final phase seamlessly supports model deployment and inference on new data, enabling users to apply their trained models to real-world tasks and make predictions effortlessly. 

\section{Experiments}



\subsection{Early Forest Fire Data Generation}
\label{sec:early-forest-fire-recognition}






Forest fires devastate the environment, human health, and ecosystems, causing extensive damage like ecosystem degradation, soil erosion, and air pollution. 
An early forest fire alert system is crucial for swift control and mitigation, but collecting quality images of early fires for training deep models is challenging. These images should be captured from devices such as drones when the fire intensity is at its lowest, resembling minimal smoke or a small fire.
FLAME dataset~\cite{flame}, derived from prescribed pile burns in Northern Arizona, 
is 
popular for this task. Although images are valuable, their creation is \textit{time-consuming, costly, and poses potential natural risks}. They lack diversity in forest types and natural conditions.
On the other hand, the images of forest fires available on the Internet are predominantly captured when the fires have spread widely.
Meanwhile, our AIR-Gen shows promise in overcoming these hurdles.

\begin{figure}[t!]
	\centering
        \includegraphics[width=0.9\linewidth]
 {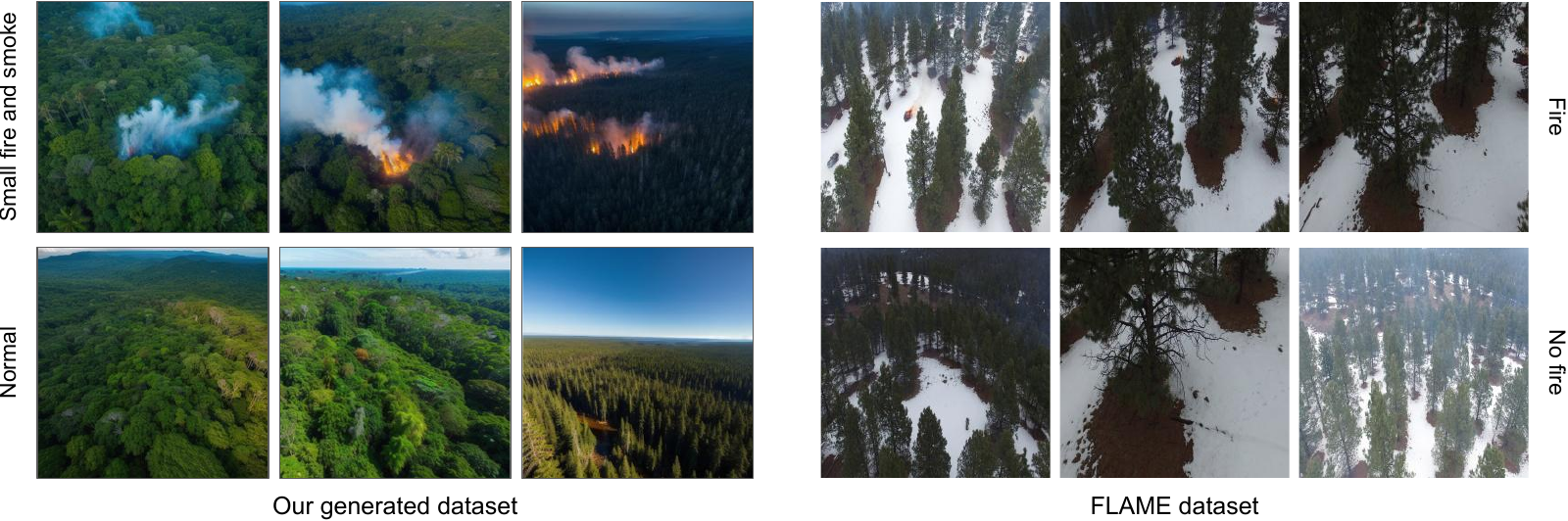}
 \vspace{-3mm}
	\caption{Samples from our generated dataset and the FLAME dataset~\cite{flame} for the task early forest fire recognition. Our dataset encompasses a diverse range of forest types and natural conditions.}
	\label{fig:generated-examples}
 \vspace{-5mm}
\end{figure}



\subsubsection{Dataset Generation}

\begin{figure}[t!]
 \begin{subfigure}{0.50\textwidth}
 \raggedright
	\includegraphics[width=\linewidth]{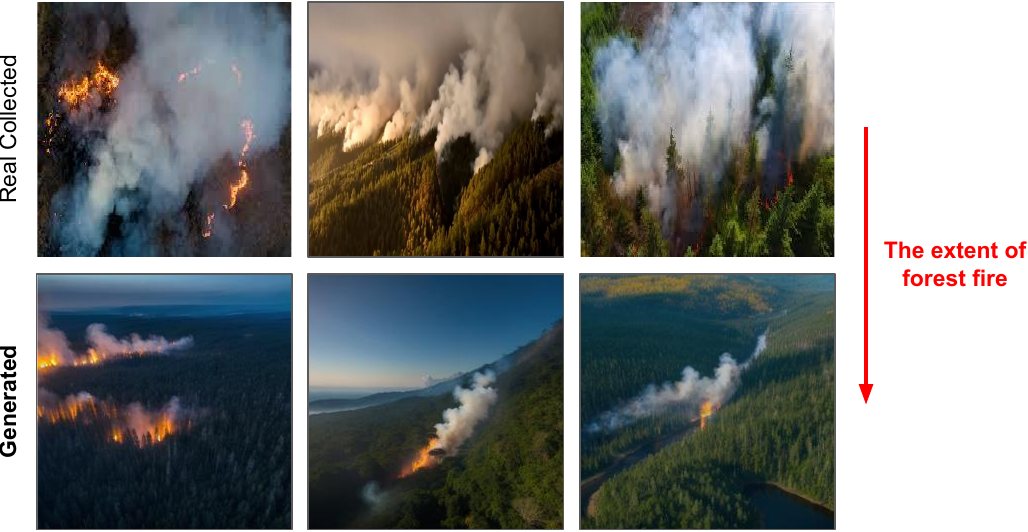}
 \caption{Image visualization}
 \label{fig:comparison-real-generated}
 \end{subfigure}
  \begin{subfigure}{0.50\textwidth}
  \raggedleft
	\includegraphics[width=0.8\linewidth]{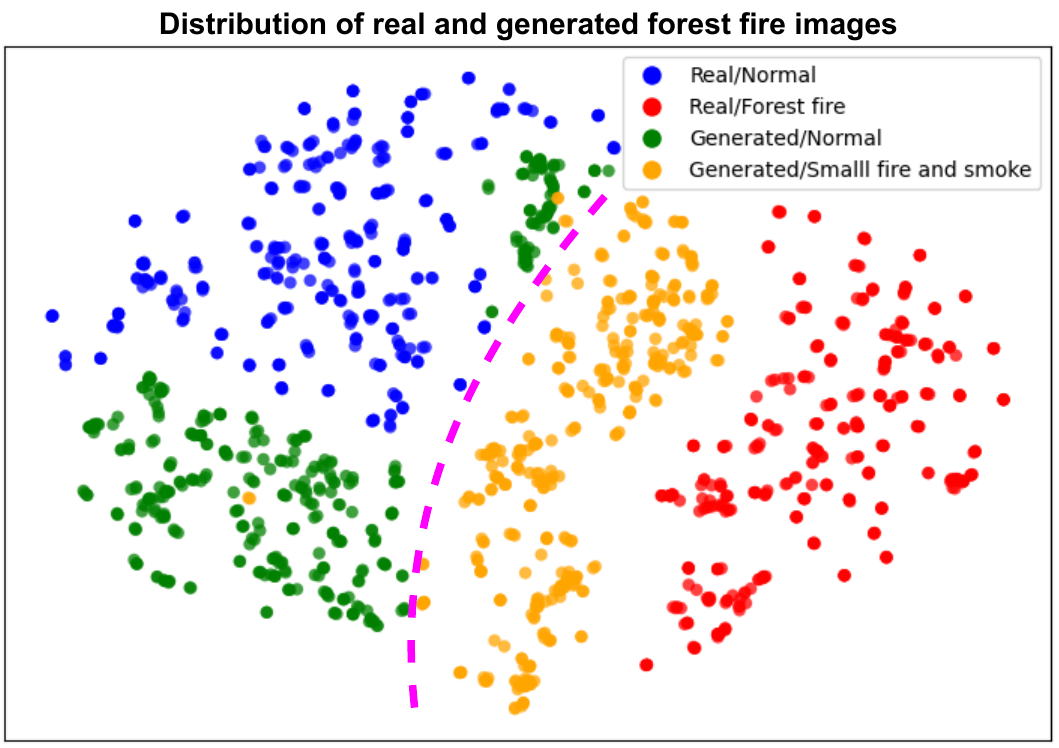}
 \caption{Feature space visualization}
 \label{fig:feature-space}
 \end{subfigure}
 \vspace{-5mm}
	\caption{Comparison of real images collected from the Internet about forest fire (which mostly depict fires that have already spread widely) and our generated images for early forest fire.}
 \vspace{-5mm}
\end{figure}

To accurately replicate drone-captured perspectives of early forest fires, we utilized AIR-Gen to create a dataset encompassing two distinct categories: ``small fire and smoke'' (4494 images) and ``normal'' (4361 images). This dataset offers a substantial volume of data for robust training and evaluation. To ensure the dataset's comprehensiveness, we generate images aligning with the variety of context settings as detailed in Table~\ref{tab:prompt_example}. 
Figure~\ref{fig:generated-examples} showcases the diversity and realism of the generated dataset compared to the FLAME dataset~\cite{flame}. 
It is noteworthy that we classify ``small fire and smoke'' as warning situations. Meanwhile, the FLAME dataset only differentiates between ``fire'' and ``no fire'', which results in images with smoke not being marked as warning situations in their dataset. Additionally, Figure~\ref{fig:comparison-real-generated} compares generated images to actual forest fire images collected from the Internet by ourselves, emphasizing a notable distinction that arises in the representation of ``small fire and smoke", a crucial indicator for early warning.





To validate the fidelity of the generated images in our dataset, we conducted a feature space visualization involving four distinct image collections: generated early forest fire, generated normal forest, real collected forest fire, and real collected normal forest. Employing CLIP image features~\cite{clip} 
for visualization, Figure~\ref{fig:feature-space} reveals two distinct clusters: normal forest and forest fire. Real and generated images within the normal forest cluster nearly overlap, indicating high resemblance. Conversely, forest fire images, both real (widely spead fire) and generated (early fire), differ considerably. The generated early fire images align more closely with the normal forest than the real forest fire images, emphasizing their dissimilarity to extensive fire images. This visualization underscores the quality and suitability of our generated dataset for training a model for the task early forest fire recognition.

We further utilize the deep model management in AIR to train classification models on our generated dataset (The training details are presented in the supplementary). To rigorously evaluate the model's performance and its ability to generalize across different subsets of the dataset, we implemented $k$-fold cross-validation with $k = 5$. Our model's performance was assessed using various metrics, including weighted accuracy, precision, recall, and F1 score. Table ~\ref{tab:result_training_app_1} summarizes our training results from each fold, indicating strong overall performance, achieving a mean accuracy of 95.48\%.

\begin{table}[t!]
\centering
\caption{Results for the task early forest fire recognition using 5-fold protocol.}
\label{tab:result_training_app_1}
\resizebox{0.75\linewidth}{!}{
\begin{tabular}{c|cccc}
\toprule
\textbf{Fold} & \textbf{Accuracy (\%)} & \textbf{Precision (\%)} & \textbf{Recall (\%)} & \textbf{F1 Score (\%)} \\
\midrule
1 & 95.49 & 95.58 & 95.49 & 95.48 \\
2 & 96.61 & 96.70 & 96.61 & 96.61 \\
3 & 95.32 & 95.42 & 95.32 & 95.32 \\
4 & 95.49 & 95.61 & 95.49 & 95.48 \\
5 & 94.51 & 94.66 & 94.51 & 94.51 \\
\rowcolor{lightgray} Mean & 95.48 & 95.59 & 95.48 & 95.48 \\
\bottomrule
\end{tabular}
}
\vspace{-5mm}
\end{table}

\subsection{MIT Indoor Scene Dataset Augmentation}









\begin{figure*}[t!]
    \centering
\includegraphics[width=\linewidth]{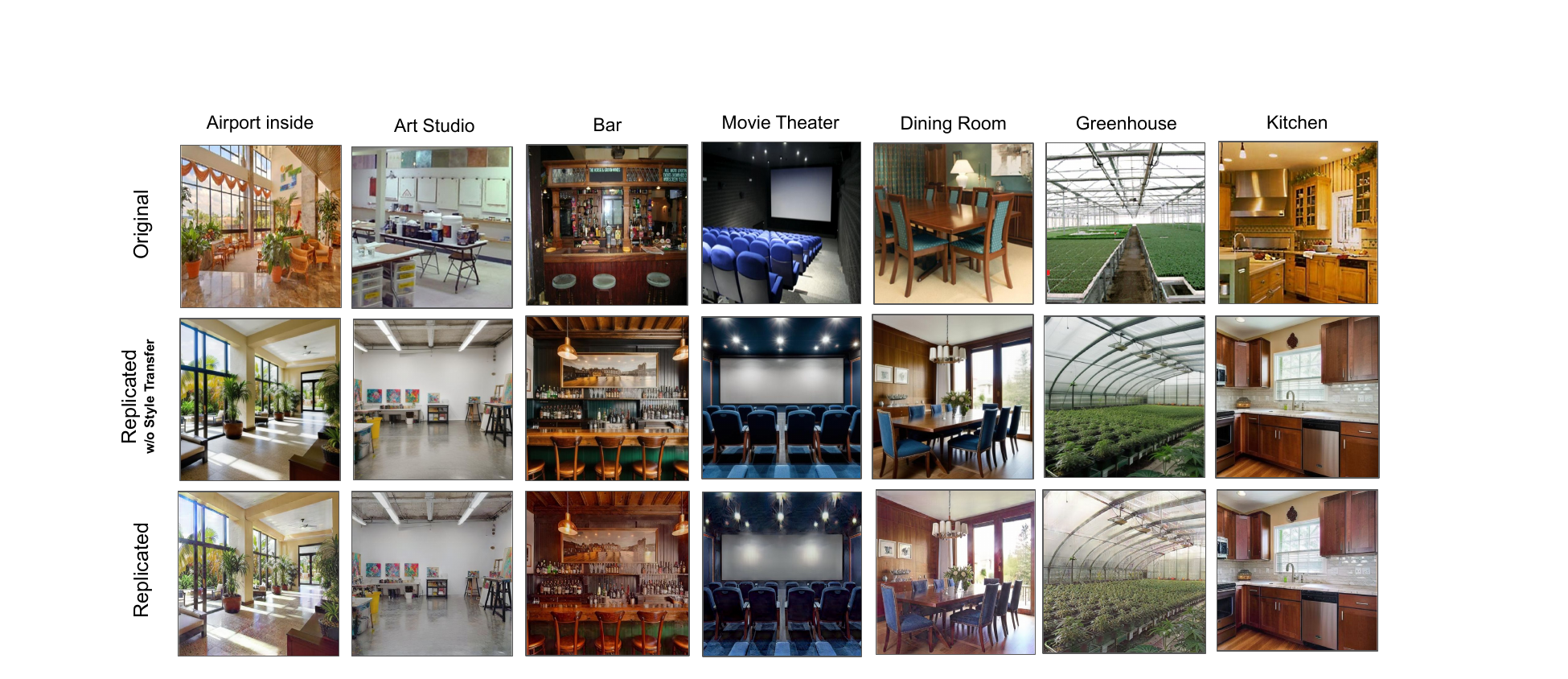}
\vspace{-5mm}
    \caption{Examples of original MIT-67 dataset~\cite{quattoni2009recognizing} (top), replicated results without Style Transfer (middle), and our replicated results (bottom).}
    \label{fig:simulated-dataset}
    \vspace{-3mm}
\end{figure*}

This experiment aims to utilize AIR-Aug to simulate a well-known dataset for image recognition, MIT Indoor Scene dataset~\cite{quattoni2009recognizing} (MIT-67 in short), then leveraging generated images as valuable augmented data to enhance deep model training, in conjunction with the original dataset. 
Through the full data synthesis process of AIR-Aug, we obtain a replicated dataset derived from the MIT-67 dataset. All replicated images are synthesized to be of size $512 \times 512$, mirroring the number of images present in the original training set. Figure~\ref{fig:simulated-dataset} presents a comparison of images from the MIT-67 dataset and their corresponding replicated versions. 
Our primary focus in the evaluation is training EfficientNetV2~\cite{tan2021efficientnetv2} models, pre-trained on ImageNet~\cite{deng2009imagenet} and fine-tuned for 25 epochs, on different training image sets: only real images, hybrid images with both replicated images and real images. We used a learning rate of $10^{-3}$, a batch size of 64, the AdamW optimizer
, and employed Sparse Categorical Cross-entropy as the loss function.

\begin{figure*}[t!]
    \centering
\includegraphics[width=\linewidth]{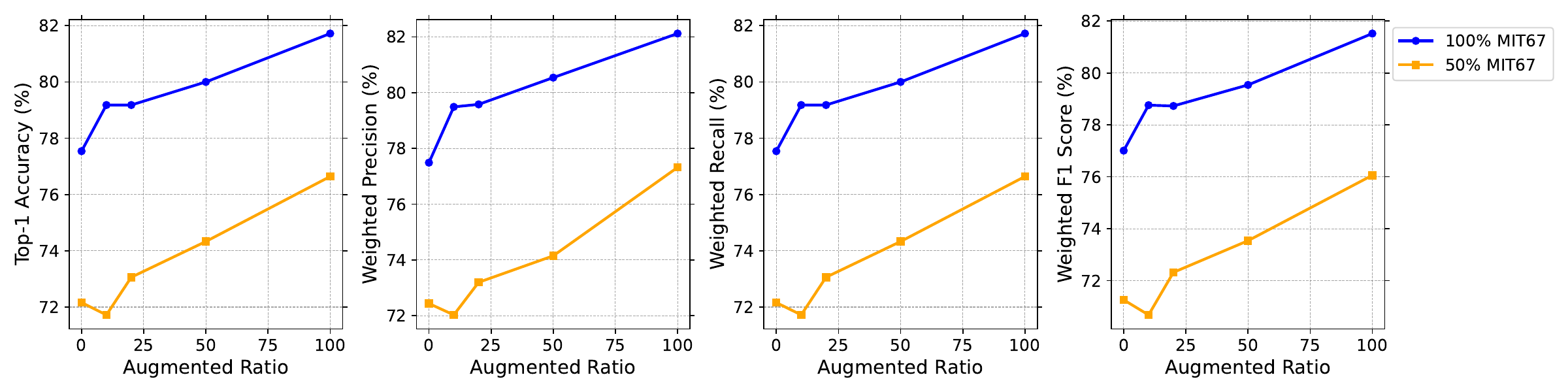}
\vspace{-5mm}
    \caption{Evaluation results of augmenting  dataset~\cite{quattoni2009recognizing} with AIR-Aug on the original test set of MIT-67. The blue lines and the orange lines respectively indicate the utilization of 100\% and 50\% number of images of the original train set to perform the augmentation.}
    \label{fig:evaluated-results}
    \vspace{-3mm}
\end{figure*}

Our evaluation process entails experiments to investigate how the quantity of replicated datasets used as augmented data influences the performance of classification metrics on the original test set of MIT-67. Regarding the combination of the original dataset and replicated dataset, we consider two options for the original dataset: utilizing 50\% of its dataset and utilizing the entire 100\% dataset. We note that using 50\% of the dataset aims to simulate the scenario where the original dataset is limited and difficult to collect. For the augmented dataset, we explore four options: 10\%, 20\%, 50\%, and 100\%. Figure~\ref{fig:evaluated-results} shows the potential benefits of including generated data from the AIR-Aug module in the training process. Particularly, we can boost the accuracy metric on the test set by roughly 4\% when the augmented dataset is fully aggregated to the original training set.


\begin{figure}[t!]

\begin{subfigure}{0.50\textwidth}
 \raggedright
	\includegraphics[width=\linewidth]{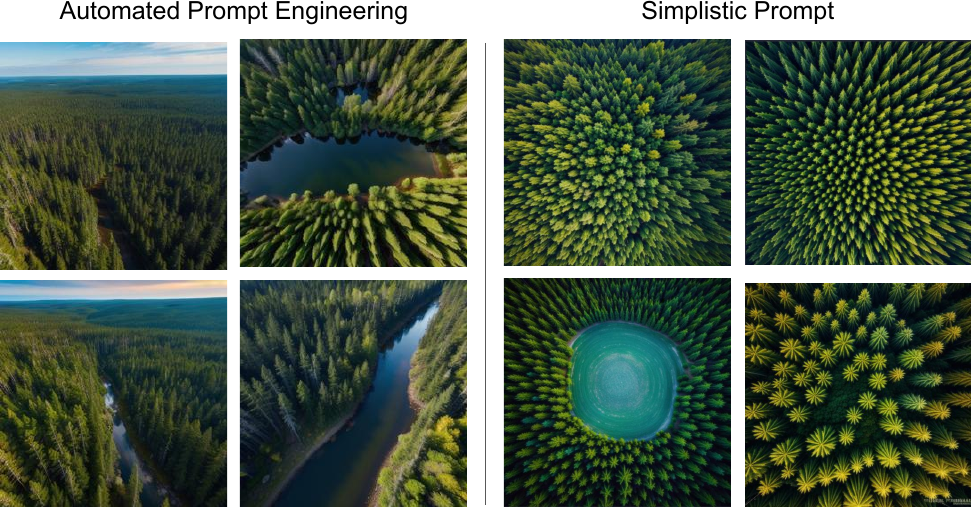}
 \caption{Image visualization}
 \label{fig:ablation_study_prompt_view}
 \end{subfigure}
  \begin{subfigure}{0.50\textwidth}
  \raggedleft
	\includegraphics[width=0.8\linewidth]{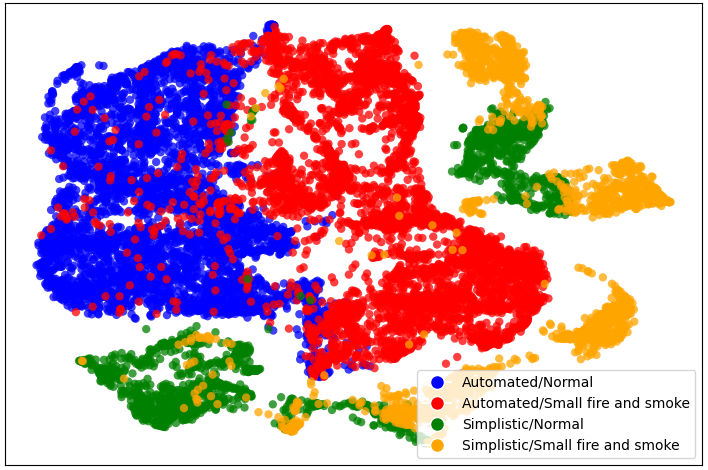}
 \caption{Feature space visualization}
 \label{fig:ablation_study_prompt_dist}
 \end{subfigure}

    \vspace{-3mm}
	\caption{Visual comparison of generated forest images using our proposed automated prompt engineering module compared with simplistic prompts.
 }
 \vspace{-3mm}
\end{figure}

\subsection{Ablation Study}


\textbf{Automated Prompt Engineering:} 
To demonstrate the effectiveness of our proposed automated prompt engineering compared to simplistic prompts (\ie, combinations of context keywords), we visualize the feature space of generated images with the context given in Table~\ref{tab:prompt_example}, using CLIP~\cite{clip} image features 
(see Fig.~\ref{fig:ablation_study_prompt_dist}). 
When simplistic prompts are utilized, the text-to-image model often produces visually similar images, resulting in numerous small clusters of green and yellow points in Fig.~\ref{fig:ablation_study_prompt_dist}. In contrast, our proposed automated prompt engineering module leads to a more reasonable and spread-out distribution. We also provide the example of generated images in Fig.~\ref{fig:ablation_study_prompt_view}. This enhancement can increase the usefulness of the generated dataset.






\textbf{Style Transfer:} As shown in Table~\ref{tab:ablation_result_1}, style transfer is crucial in data augmentation. Utilizing CycleGAN for style transfer can notably enhance the top-1 accuracy by 3.2\% on the fully replicated dataset, resulting in an accuracy of 81.72\%.


\begin{table}[t!]
\centering
\caption{The top-1 accuracy (\%) using the original training set of MIT combined with our augmented data on the original test set of MIT-67 between our method, without duplicated and outlier removal, and without style transfer.}
\label{tab:ablation_result_1}
\resizebox{0.8\linewidth}{!}{
\begin{tabular}{l|l|l|l|l|l}
\hline
\backslashbox{\textbf{Method}}{\textbf{Aug. (\%)}}                                    & \multicolumn{1}{|c|}{\textbf{0}}      &  \multicolumn{1}{|c|}{\textbf{10}}    & \multicolumn{1}{|c|}{\textbf{20}}    & \multicolumn{1}{|c|}{\textbf{50}}  & \multicolumn{1}{|c}{\textbf{100}}   \\ \hline
\rowcolor{lightgray} Ours                                & 77.54 & 79.18 & 79.18 & 80.00  & 81.72 \\
w/o Removal &    --   & 77.15 \textcolor{red}{(-2.03)} & 78.58 \textcolor{red}{(-0.6)}& 79.10 \textcolor{red}{(-0.9)} & 80.00 \textcolor{red}{(-1.72)}   \\
w/o Style Transfer                 &   --    &    77.83 \textcolor{red}{(-1.35)}  &  77.76 \textcolor{red}{(-1.42)}     &    78.06 \textcolor{red}{(-1.94)} &   78.50 \textcolor{red}{(-3.22)}    \\ \hline
\end{tabular}
}
\vspace{-3mm}
\end{table}


\begin{wrapfigure}{r}{0.50\textwidth}
  \begin{center}
    \includegraphics[width=0.48\textwidth]{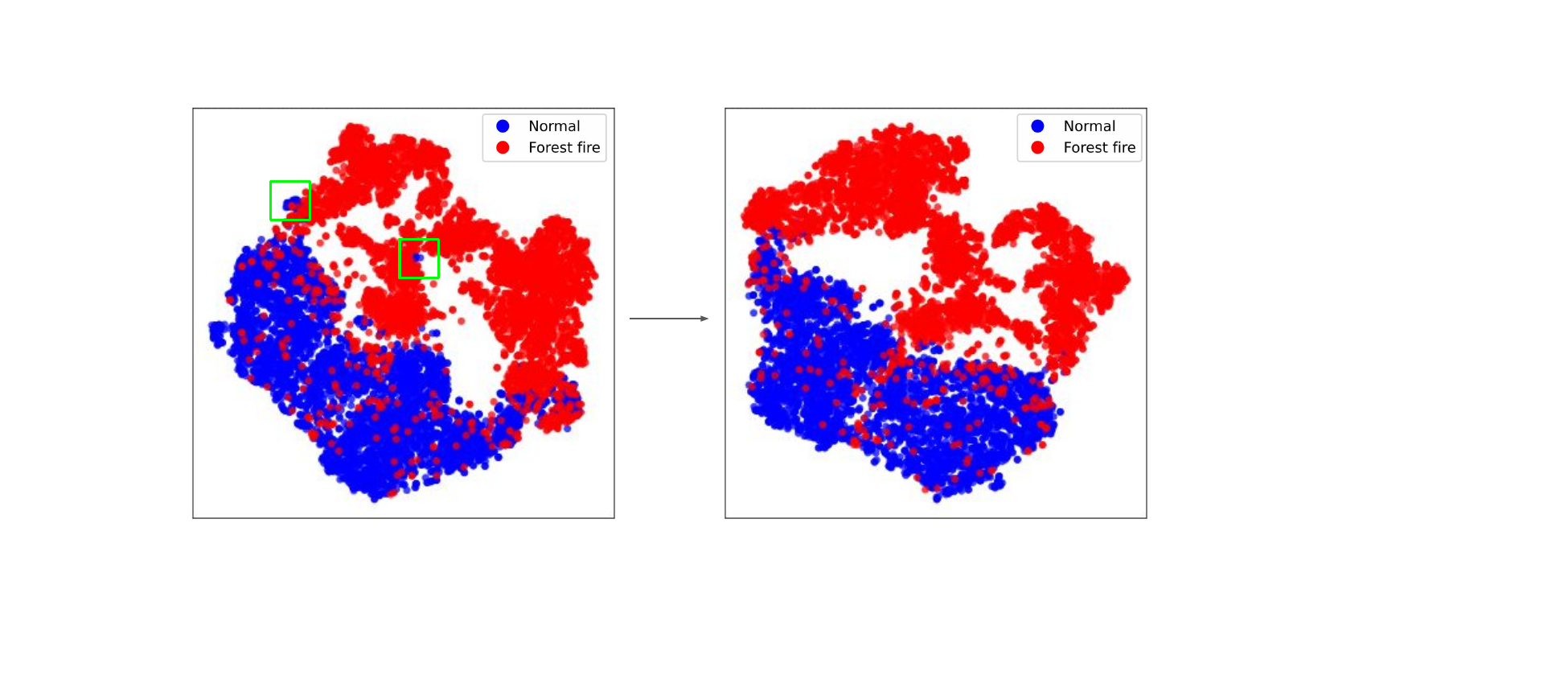}
  \end{center}
  \vspace{-5mm}
  \caption{Feature space visualization of our generated images before and after using duplicated and outlier removal.}
    \label{fig:ablation_study_filter_removal}
\end{wrapfigure}

\textbf{Duplicated and Outlier Removal:} Regarding AIR-Gen, Figure~\ref{fig:ablation_study_filter_removal} visualizes the feature space before and after eliminating duplicates and outliers with the context provided in Table~\ref{tab:prompt_example}. This process eliminates some local clusters of the "Normal" class that were too far away from the major cluster. Meanwhile, for AIR-Aug, we conduct experiments to determine the effectiveness of our module in improving the quality of the replicated dataset used for data augmentation. Table~\ref{tab:ablation_result_1} indicate that the removal module can improve the top-1 accuracy by 1.72\%, achieving an accuracy of 81.72\% on the fully replicated dataset.


\subsection{User Study}

We conducted a user study to evaluate the effectiveness of AIR, especially AIR-Gen, in supporting Computer Science (CS) students and experts in their projects. Our study involved 20 participants, of which half of them (55\%) were CS undergraduates with limited AI expertise, still in the process of early learning AI. The remaining participants consisted of those with actual AI experience by practically working in the field, including CS graduate students (15\%), AI researchers, and AI engineers (30\%).

We launched AIR and invited participants to experience and rate its performance across various aspects.
Our study primarily aims to measure users’ satisfaction in five criteria, namely, ease of usage, synthesized image quality, dataset usefulness, trained model performance and reproducibility portability. 
Table~\ref{tab:survey_result} summarizes the Mean Opinion Score (MOS) and the Median Interquartile Range (IQR) of each factor on a scale of 1 to 5 (1: least satisfied, 5: most satisfied) from the user study evaluation. Our system obtains an impressive overall score of 4.4 out of 5.0, reflecting highly positive overall user experiences. 

By enabling users to download their trained models and offering an easy-to-use Google Colab running script, we achieve a high score on the reproducibility of model predictions. However, our users often request datasets focused on very specific topics, such as anime, game styles, or landmarks in certain countries. These preferences might not match the strengths of the Realistic Vision checkpoint used in our Stable Diffusion model for image generation. This mismatch results in lower scores for image photorealism and the alignment of the generated dataset with their expectations.

\begin{table}[t!]
\centering
\caption{Summary of user study results, using Mean Opinion Score (MOS) and Median Interquartile Range (IQR). Scores range from 1 (least satisfied) to 5 (most satisfied).}
\label{tab:survey_result}
\resizebox{0.85\linewidth}{!}{
\begin{tabular}{l|c|c}
\toprule
\textbf{Evaluated Features} & \textbf{MOS} & \textbf{Median (IQR)} \\
\midrule
Ease of interaction even without AI familiarity & 4.20 & 4 (1) \\
Photorealistic image generation & 3.95 & 4 (0.5) \\
Alignment with dataset expectations & 3.95 & 4 (0.5) \\
Trained models performance & 4.40 & 5 (1) \\
Reproducibility of model's predictions & 4.75 & 5 (1) \\
\rowcolor{lightgray} Overall & 4.40 & - \\
\bottomrule
\end{tabular}}
\vspace{-3mm}
\end{table}

\section{Conclusion and Future Work}

In this paper, we introduced AIR, a comprehensive solution tailored for automated image recognition. 
Our proposed method comprises a data generation pipeline for crafting customized high-quality datasets and a data augmentation process to further enhance given datasets effectively, alongside a complete framework for training, evaluating, and inferring image recognition models. 
Through extensive experiments, we demonstrated the effectiveness of our generated and augmented data in training deep learning models, and showcased the system's capability to produce image recognition models for a wide variety of concepts.
The user study reveals that AIR can empower a broad spectrum of users, including those without extensive expertise in AI.

It is also noteworthy that the data generation process of AIR is still limited by the capability of the Stable Diffusion model. Consequently, AIR does not perform well in certain specific domains, such as medical image or non-realistic image synthesis. To broaden AIR's applications, we aim to expand its potential by incorporating more Stable Diffusion checkpoints and automatically switching between them to optimize performance for specific dataset requirements. 

\textbf{Acknowledgments. } 
 This research is funded by Vietnam National University - Ho Chi Minh City (VNU-HCM) under Grant Number C2024-18-25.













%
%
\bibliographystyle{splncs04}
\bibliography{main}

\end{document}